\documentclass{rob}%

\usepackage{amssymb}
\usepackage{amsmath}
\usepackage[super]{cite}
\usepackage{graphicx}
\usepackage{subcaption,xcolor,hyperref}

\doi{}
\volume{}
\issue{}
\pubyear{2021}
\cpyear{}
\endpage{}

\begin{document}

\title{Design and Integration of a Drone based Passive Manipulator for Capturing Flying Targets\vspace{1cm}}
\author{Vidyadhara B. V.$^{1}$\thanks{Corresponding author. E-mail:
vidyadhara.vk@gmail.com}, Lima Agnel Tony$^{1}$\thanks{Corresponding author. E-mail:
limatony@iisc.ac.in}, Mohitvishnu S. Gadde$^{1}$, Shuvrangshu Jana$^{1}$,
 Varun V. P.$^{2}$, Aashay Anil Bhise$^{1}$, Suresh Sundaram$^{3}$, Debasish Ghose$^{1}$}\affil{$^{1}$Guidance, Control, and Decision Systems Laboratory (GCDSL), Department of Aerospace Engineering, Indian Institute of Science, Bangalore-12, India.\\$^{2}$Robert Bosch Center for Cyber Physical Systems, Bangalore-12, India.\\$^{3}$ Artificial Intelligence and Robotics Laboratory (AIRL), Department of Aerospace Engineering, Indian Institute of Science, Bangalore-12, India.\\Emails: mohitvishnug@iisc.ac.in, shuvrangshuj@iisc.ac.in, varunvp@iisc.ac.in, meetaashay3@gmail.com, vssuresh@iisc.ac.in, dghose@iisc.ac.in}
\ADaccepted{MONTH DAY, YEAR. First published online: MONTH DAY, YEAR}

\maketitle

\begin{summary}
In this paper, we present a novel passive single Degree-of-Freedom (DoF) manipulator design and its integration on an autonomous drone to capture a moving target. The end-effector is designed to be passive, to disengage the moving target from a flying UAV and capture it efficiently in the presence of disturbances, with minimal energy usage. It is also designed to handle target sway and the effect of downwash. The passive manipulator is integrated with the drone through a single Degree of Freedom (DoF) arm, and experiments are carried out in an outdoor environment. The rack-and-pinion mechanism incorporated for this manipulator ensures safety by extending the manipulator beyond the body of the drone to capture the target.  The autonomous capturing experiments are conducted using a red ball hanging from a stationary drone and subsequently from a moving drone. The experiments show that the manipulator captures the target with a success rate of 70\% even under environmental/measurement uncertainties and errors.
\end{summary}

\section{Introduction}\label{s0}
Technological advancements in Unmanned Aerial Vehicles (UAVs) have led to the growth of industries developing solutions for various civilian and military applications. UAVs are extensively used for various applications like aerial photography, package delivery, mapping of difficult terrain or environments, reforestation, visual inspection, search and rescue operations \cite{srch&res}, etc. Drone delivery networks like Amazon Prime Air \cite{amazon}, UAV based medical transportation like Zipline \cite{zipline}, drone photography with DJI \cite{dji}, Skydio \cite{skydio}  are platforms of a few popular application. Due to their complexity, UAVs and their related sub-systems pose a challenge to researchers. Recently, aerial manipulation is gaining attention due to its wide scope for applications. Manipulation mechanisms are key to robotics applications. An ideal manipulator should be energy optimal and have low response time and structural integrity. Fields in which this domain will have an impact include material handling, inventory management, package delivery, etc. Steps towards achieving similar tasks are available in the literature.

The literature has diverse works on aerial manipulation \cite{8299552,meng2020survey} for different applications. Controlling a valve\cite{6842330} using multiple DoF manipulator and adopting parallel manipulators for turning\cite{Rob1} are those which are suitable for localised manipulation with higher accuracy. Tri-finger end effector design is adopted in \cite{origami}, which is a foldable one and utilises less power. Manipulation using an industrial manipulator on a helicopter \cite{6907148} is also adopted where, a seven DoF manipulator is employed to study the coupling effects of the integrated system. This manipulator has better maneuverability but the end-effector operational area is small with limited reach and considerable weight. 
Only a few research papers\cite{6875943,7158852} in the literature discuss the design and modeling of aerial manipulators. Pick and place operation using haptic control\cite{8621786} of manipulators are looked into, where the end-effectors are inefficient for a dynamic task and have limited operational volume. Manipulators for cooperative transportation \cite{8120115,tagliabue2017collaborative,gioioso2014flying} of objects work together to achieve static object transportation. Manipulators for contact-based operations are also actively researched \cite{hamaza2018adaptive,8206398} for problems like pipeline and power line monitoring and repair and similar applications. 
\begin{figure}[h]
    \centering
    \includegraphics[scale=0.35]{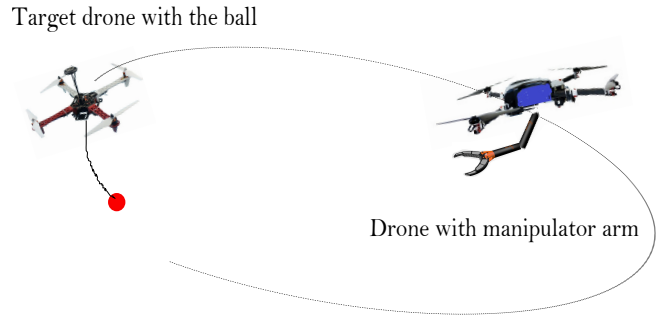}
    \caption{Sample scenario describing the problem}
    \label{fig:prob_def}
\end{figure}
Aerial grasping of objects is another interesting area. A vision-based grasping is presented in \cite{7395364}, where a multi-degree of freedom robotic arm is considered. Aerial manipulation of a rod-shaped object\cite{8253830} using multiple robots is another stationary object manipulation in which, the manipulator's task is to grip the longitudinally placed object. Grasping of cylindrical objects is presented in \cite{7989751} while a suction-based end effector for pick and place is given in \cite{7487495,8793672}. An extending zipper manipulator for aerial grasping is presented in \cite{8967982}, while a seven DoF manipulator with a hex-rotor\cite{8461103} for object grasping. 

The above-mentioned designs are used to interact with static targets. They mainly incorporate multi DoF robotic arm concepts for the task. These designs are heavy, have high power consumption and add computational and mechanical complexity. For the considered task, which is extremely dynamic and fast, these are major issues. While the downwash and vibration aspects are not serious concerns in the reviewed literature, it is a serious problem for a dynamic target capture task. Most of the works that deal with moving object grasping are dealt with in a control perspective rather than a manipulator design perspective. Hence, a novel design that can interact with dynamic targets and accommodate considerable sensor information error is a much-needed research and has applications in several domains. 

In this paper, we present modelling and development of a single DoF manipulator for aerial grabbing of stationary and moving objects in an outdoor environment. The design contributes to low drag and low impact from downwash. The proposed passive end-effector design is energy optimal and any object within 0.15 kg and 0.2 m diameter could be grabbed. The design of the manipulator is presented along with the analysis on the manipulator modelling parameters and the stability of the integrated system. The paper also presents results demonstrating the performance of the manipulator while grabbing stationary and moving object. 

The paper is organized as follows: Section \ref{sec:2} describes the problem statement along with the design requirements. Section \ref{sec:3} presents the challenges involved in achieving the task. Section \ref{sec:4} gives the detailed design of the passive manipulator, and the material used for the prototype. Section \ref{sec:5} gives the design and structural analysis of the integrated system. The experimental setup, including the drones, avionics, and the test results are provided in Section \ref{sec:6}. Section \ref{sec:7} concludes the paper.
\section{Problem Description}\label{sec:2}
The manipulator is designed for aerial grabbing of moving targets. The problem represented in Fig. \ref{fig:prob_def},
is inspired by Challenge 1 of MBZIRC 2020 \cite{MBZ}, in which a drone carries a ball attached to it with a flexible rod. The drone moves with a maximum speed of $6$ m/s, and the ball weighs $0.060$ kg and is $0.15$ m in diameter. The contact between the rod and the ball is magnetic. The task is considered successful if the drone can detach the ball and drop the captured ball in a box. 
The ball is prone to oscillations because of the drone maneuvers and environmental factors like wind gusts and downwash. The design requirements for the manipulation mechanism are listed below.

\renewcommand{\theenumi}{\alph{enumi}}%
\begin{enumerate}
    \item The volume of the integrated system should be within $1.2$ m $\times$ $1.2$ m $\times$ $0.5$ m during take-off and landing.
    \item The manipulator should be able to exert the detachment force of $4$ N while maintaining structural integrity.
    \item The captured ball should be deposited in a box, and hence requires a release mechanism.
    \item The ball sways because of the motion of the  drone and due to external disturbances. The design should be able to handle the uncertainties in position of the ball.
  \item The integrated system should be stable to capture the maneuvering target.
\end{enumerate}
 
\section{Manipulator Design Challenges}\label{sec:3}
The design of the manipulator and the end-effector should address these challenges for effective grasping.
    \begin{figure}
	\centering
	\includegraphics[scale=0.35]{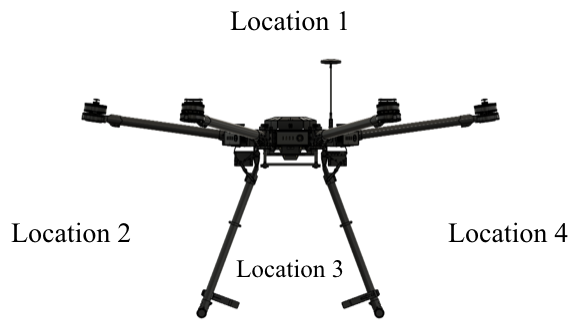}
    \caption{Locations where an end-effector for grabbing can be mounted on a UAV}
    \label{fig:location}
    \end{figure}
\renewcommand{\theenumi}{\alph{enumi}}%
\begin{enumerate}
    \item Location of the manipulator: Selecting the location to mount the manipulator is crucial. Fig. \ref{fig:location} represents the possible location of end-effector. Location 1 has a large usable volume. Location 1 and location 3 have the advantage that they can be placed very close to the center of drone's frame but pose the risk of the drone or the ball striking the propellers. Metallic construction around the GPS can cause problems, making location 1 a risky choice. Locations 2 and 4 have large usable volumes but are affected by the downwash from the propellers. Locations 2, 3, and 4 require the manipulator to extend away from the drone body to ensure safety and avoid downwash. It may generate moments about the drone centre of gravity (CG) if it is not properly stabilised.
    \item Length of the extension arm: The end-effector should be at a safe distance from the propellers, to ensure safety. This is done by extending the end-effector away from the drone body, as shown in Fig. \ref{fig:extension} using different mechanisms. It could be a single DoF (A/B), two DoF (C) or multiple DoF (D) manipulation mechanisms. Multiple DoF improves the reachable space of the manipulator at the cost of increased computational and control complexity.
    Having multiple DoF adds to the reachable space of the manipulator but at the cost of increased computational and control complexity. 
    While a considerable extension of the arm is recommended for the safety of the drone, it might contribute to several other issues. 
    Large distances from the point of attachment create a noticeable deflection at the end-effector. For multiple DoF joints, this affects the performance of actuators resulting in sensor errors due to change in orientation and position. Such extensions also create moments that tend to destabilize the drone. 
    \begin{figure}
        \centering
        \includegraphics[scale=0.4]{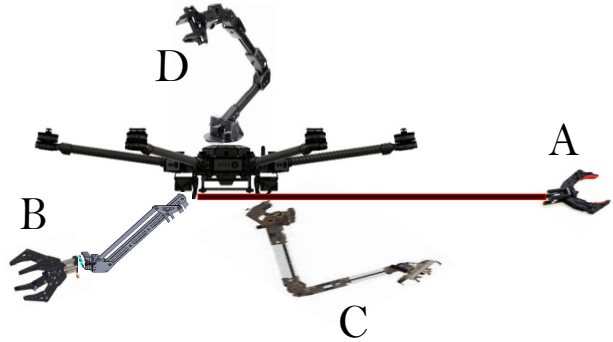}
        \caption{Manipulator arm design considerations}
        \label{fig:extension}
\end{figure} 
    \item Detachment force: The forces encountered while detaching a ball from the target UAV are shown in Fig. \ref{fig:detach}. The ball is attached to the rod suspended from the UAV air frame, by a magnet. The detachment force (denoted by $F_d$ in Fig. \ref{fig:detach}) is non-uniform as it depends on the way that the magnets are pulled apart, and thus depends on the mechanism of grasping.
    \begin{figure}
        \centering
        \includegraphics[scale=0.3]{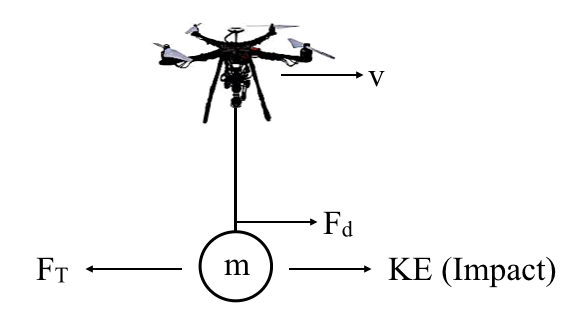}
        \caption{Requirement of a force to remove target}
        \label{fig:detach}
    \end{figure}
    In addition, the target UAV is moving at a velocity $v$, which causes a force on the end-effector as a result of the kinetic energy acquired by the ball. The end-effector must be able to exert the maximum necessary force. Creating a robust manipulator with a high factor of safety, results in an increase in weight adding to the moment and sag issues. This challenge relies on strength to weight optimisation and material selection. 
    \item Vibrations: The end-effector self-weight causes deflection, as mentioned above. This deflection imparts vibrations. The manipulator along with the end-effector acts as an end loaded cantilever beam which is  prone to vibrations even for small disturbances at the free end. Hence, sensor vibration dampening becomes crucial for this task, to remove noise from the vision feed. 
\end{enumerate}

\section{Passive Manipulator Design Approach}\label{sec:4}
The challenges involved in the problem are thoroughly examined and the following design is proposed. The design requires certain essential capabilities like quick response, low weight, and optimal power consumption. The design is a result of iterative and progressive developments from a preliminary concept \cite{arxived}.

\subsection{Passive end-effector}
\begin{figure}
	\centering
    \includegraphics[scale=0.75]{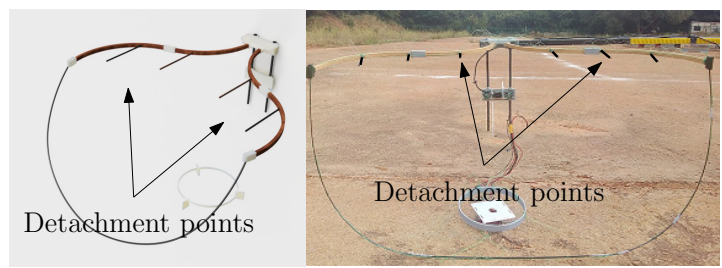}
    \caption{(a) CAD model of the final passive ball grabbing end-effector (b) A working prototype }
    \label{fig:c1_EF5}
\end{figure}
\begin{figure}
	\centering
	\includegraphics[scale=0.75]{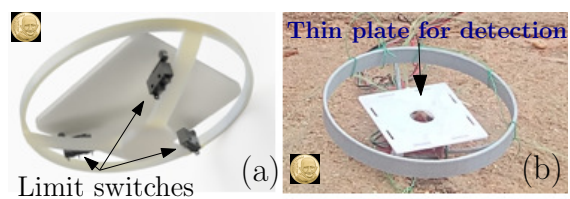}
    \caption{Grab detector (a) CAD model (b) Prototype (coin for scale)}
    \label{c1-grab-det}
\end{figure}
The motivation of the manipulator mechanism comes from the passive fruit pickers used in orchards. The design is as shown in Fig. \ref{fig:c1_EF5}(a) and the prototype of this design is shown in Fig. \ref{fig:c1_EF5}(b). As seen in the figures, the top portion of the end effector has a sinusoidal shape made from birch, with several detachment points made from carbon fibre (CF) strips. A hand woven nylon mesh is attached to its bottom and is supported at the front using a semi-circular CF ring. The specific shape of the top not only supports the mesh below but also improves its effectiveness. The top portion of the end-effector is convex at its center and concave towards the ends. The center portion of the end-effector has 3 D printed mounts to place the camera (eye-in-hand configuration) using CF tubes and to attach the basket on to the manipulator arm. The convexity ensures that the ball remains within the FoV of the camera until it is detached, which otherwise would happen to the side or behind the camera. The concave shape towards the end give sufficient room for the ball to be collected in the basket. If not, the ball may fall outside while being detached from the drone. The CF detachment points aid in effectively detaching the ball.

Since, the target needs to be dropped in a box, a dropping mechanism is integrated into the system. The servo motor present in the dropping mechanism helps in releasing the ball in the box. The servo motor requires lesser power to operate satisfying the minimal power requirement.
An additional requirement for the manipulator is to detect the grabbed ball. Approaches like visual feedback with a separate camera or gimbal mounted camera would add computational load on the system, with considerable energy requirement. Thus, a thin plate detector is designed with three switches placed around a circle, as shown in Fig. \ref{c1-grab-det}(a). The plate at its center improves sensitivity and detects the grabbed ball. 
The design uses gravity for grab detection and release of the ball into the basket, with minimal use of energy. The prototype of the mechanism, is shown in Fig. \ref{c1-grab-det}(b). 
The gray rim is held firmly by the mesh. This is the lowest end of the passive basket end-effector. 
\subsection{Manipulator Arm}
\begin{figure}
	\centering
    \includegraphics[scale=0.35]{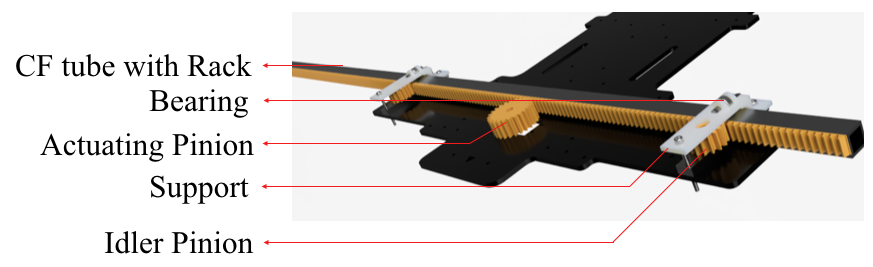}
    \caption{CAD model showing the idler pinion support assembly }
    \label{fig:idler}
\end{figure}

The development of the end-effector is also influenced by the choice of the drone. The drone selected for testing is DJI M600 \cite{M600}. It was chosen as it fits within the size constraints mentioned in Section \ref{sec:2} and also provides a flight time up to 30 minutes. 
The end-effector is designed to be positioned at the side of the drone. This ensures safe detachment of the ball, reducing the possibility of head-on crashes with the drone carrying the ball. The manipulator is extended sideways via a rack and pinion mechanism. It is desired to have minimum vibrations and play along horizontal and vertical planes. This is achieved by using idler pinion gears with bearings shown in Fig. \ref{fig:idler}. The black acrylic plate holds the manipulator and is attached to the bottom of the drone frame. The manipulator extension arm is a 15 mm $\times$ 15 mm CF square tube of 1.2 m length. 
The idler pinions provides the required tension while facilitating smooth actuation. In order to avoid vertical deflection, the rigidity of the arm is increased, which reduces the vibrations due to the deflection experienced by the arm. 

\section{Analysis}\label{sec:5}
In this section, the proposed manipulator design is analysed. The major aspects examined here are the end effector dimensions, manipulator arm extension limits, location of camera, and the structural stability of the integrated system.
\subsection{End-effector dimensions}

Considering the nature of the problem, two factors contribute to the successful grasping: grab volume and capture area. Grab volume is the effective volume available at the end-effector to capture the ball. 
Capture area is the effective area of the end-effector that engages with the ball to detach it from the target drone.
The lower bound of grab volume is defined by the size of the object to be grabbed. The volume constraint limits the maximum grab volume of the end-effector.
The capture area is upper bounded by the size constraints of the integrated system and lower bounded by the FoV constraints from the camera, which is described in the next section.
\begin{figure}
	\centering
	\includegraphics[scale=0.75]{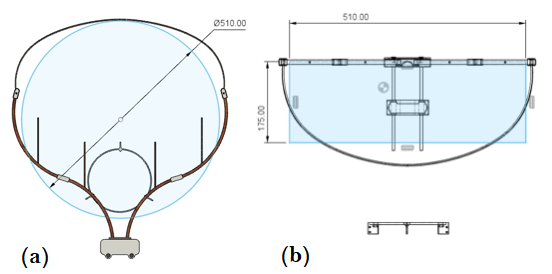}
    \caption{Regions of interest for grab volume and capture area calculations using (a) top view (b) front view of the end-effector}
    \label{c1-approx}
\end{figure} 
Based on the shape of the end-effector, a truncated cone best represents the approximate volume of the end-effector. The top and front view of the passive basket marked with respective dimensions, are shown in Fig. \ref{c1-approx}. The upper structure is approximated as a circle of diameter 0.51 m, as shown in Fig. \ref{c1-approx} (a). The capture area of the final end-effector is approximately a rectangle of sides shown in Fig. \ref{c1-approx} (b). The capture area is 
$A_{\text{cap}}^{\text{passive}}= 0.51\times0.175=89.25\times10^{-3} ~\text{m}^2$.

\begin{figure}
    \begin{subfigure}{0.45\columnwidth}
        \centering
        \includegraphics[scale=0.6]{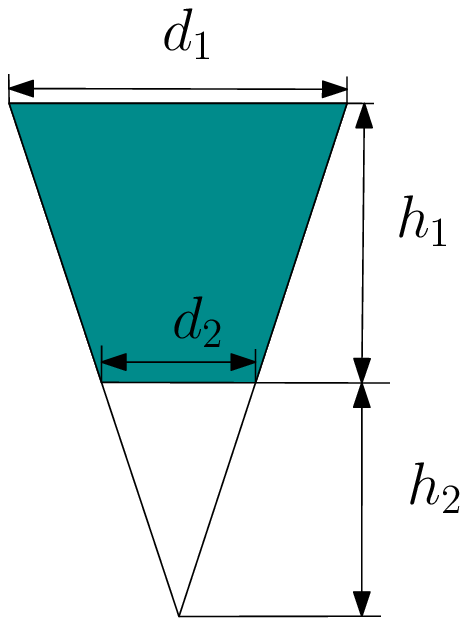}
        \caption{}
        \label{c1-Trunc-cone}
    \end{subfigure}
    \begin{subfigure}{0.45\columnwidth}
        \centering
    	\includegraphics[scale=0.45]{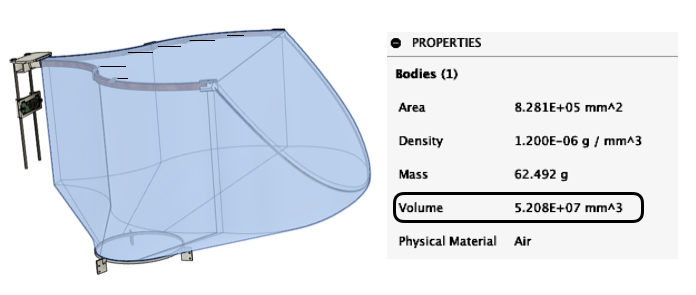}
        \caption{}
        \label{c1-Vol-Act}
    \end{subfigure}
    \caption{(a) Approximate truncated cone for volume calculation (b) Precise grab volume determined from the CAD model of the final design}
\end{figure}

A ring size of 0.175 m is found feasible for the ball detector. The truncated cone as shown in Fig. \ref{c1-Trunc-cone}, is constructed with dimensions as given in Table \ref{t1}. Hence the approximate grab volume is $V_{g_\text{approx}}^{\text{passive}}$ = 34.817$\times$10$^{-3}$ ~\text{m}$^3$
\begin{table}
    \centering
    \caption{Dimensions of the truncated cone}
    \begin{tabular}{cc}
    \hline
     Parameter    & Value (m)\\\hline
       $d_1$  & 0.51\\
       $d_2$ & 0.175 \\
       $h_1$ & 0.35\\
       $h_2$ & 0.1828\\\hline
    \end{tabular}
    \label{t1}
\end{table}
The CAD model of the basket is shown in Fig. \ref{c1-Vol-Act}. The precise grab volume is determined as $52.08\times10^{-3} ~\text{m}^3$. A 33\% increase from the approximate volume could be achieved in the final design by adjusting the mesh shape to a accommodate a larger volume, during manufacturing. A major factor for the effectiveness of the design is the large capture area and grab volume, which greatly helps in handling small disturbances and oscillations of the ball due to wind or maneuvers.
\subsection{Camera location}
\begin{figure}
    \centering
    \begin{subfigure}{0.5\columnwidth}
        \centering
        \includegraphics[scale=0.45]{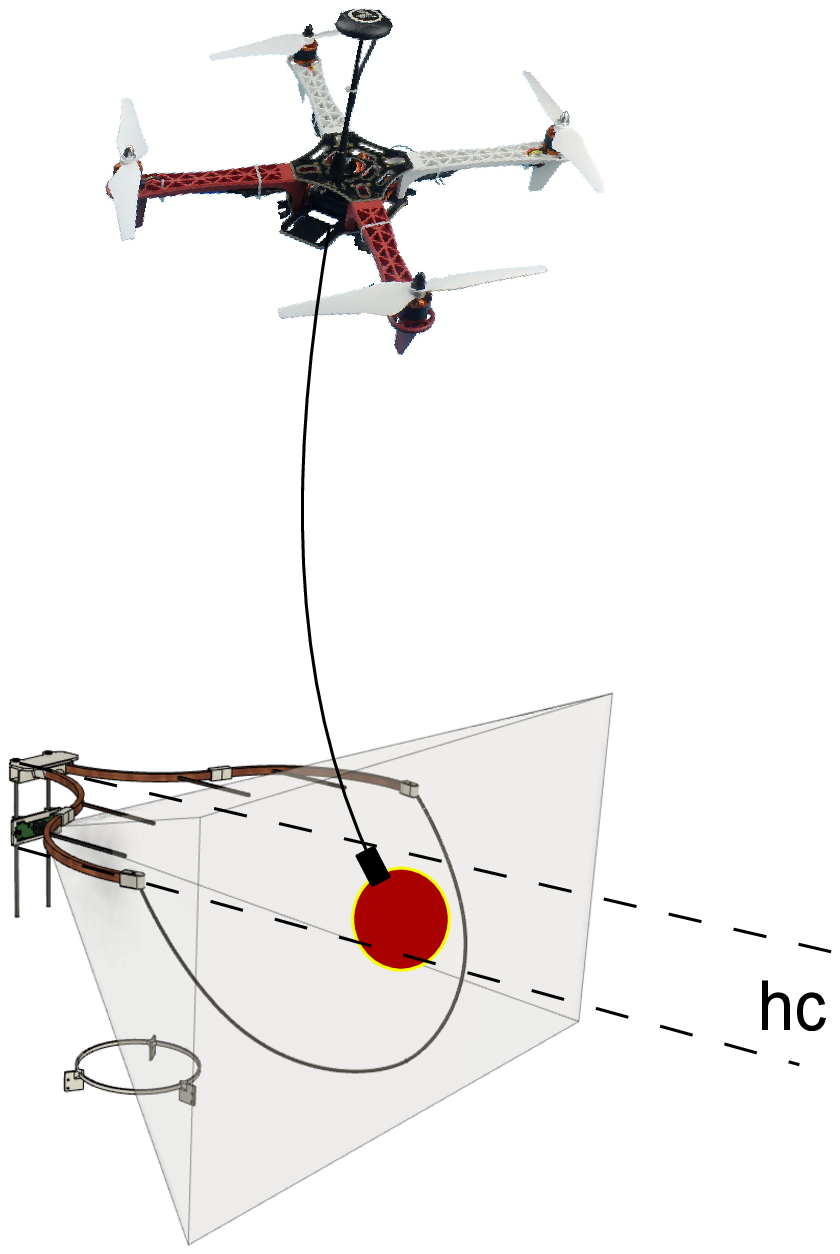}
        \subcaption{}
    \end{subfigure}
    \begin{subfigure}{0.45\columnwidth}
        \centering
        \includegraphics[scale=0.35]{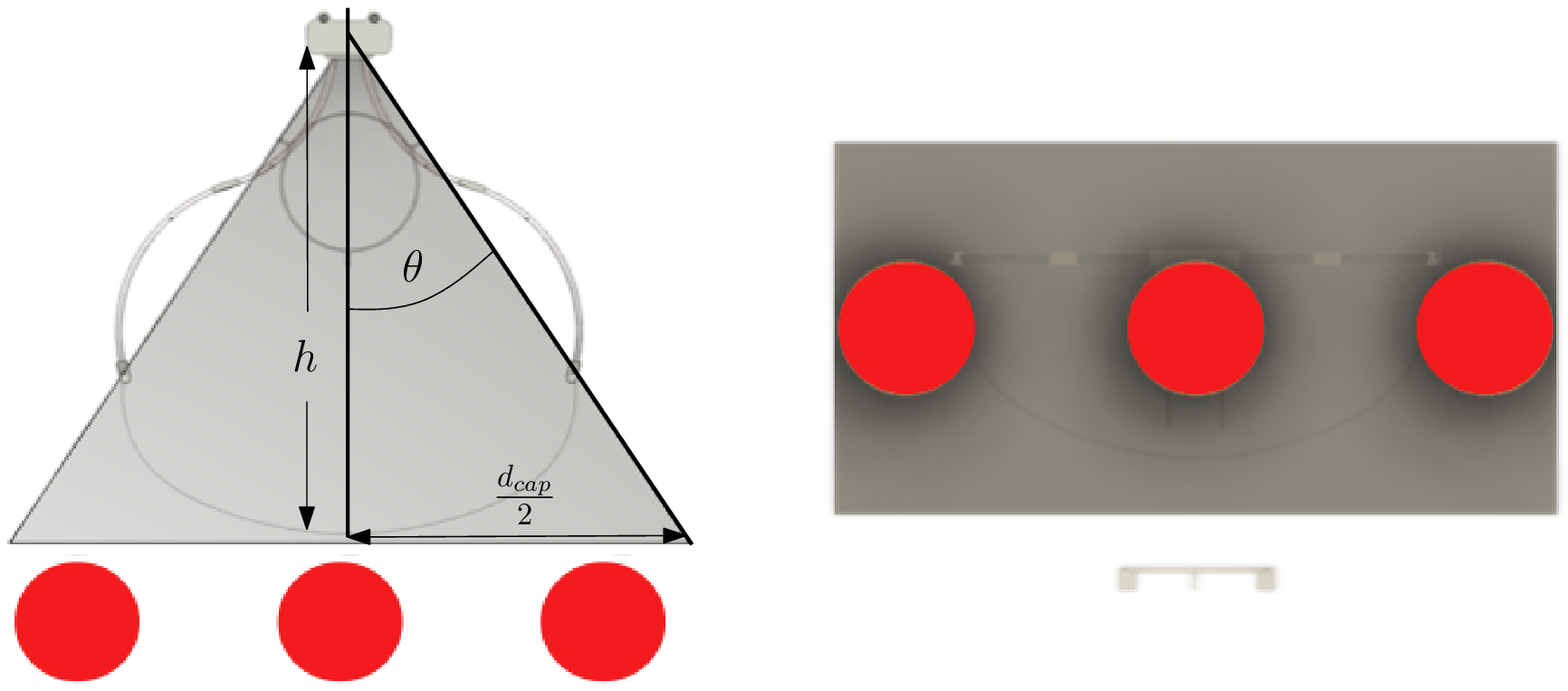}
        \subcaption{}
    \end{subfigure}
    \caption{(a) Camera positioning along the vertical plane (b) FoV considerations for deciding the basket opening}
    \label{c1-cam-loc}
\end{figure}

As described in Section \ref{sec:4}, an eye-in-hand configuration is ideal for the proposed design. The location of the vision sensor is important for two reasons. The vertical placement of the camera should be such that the camera center and the ball center should coincide and the ball should be within the basket. That is, if $h_c$ is the location of the camera below the basket top and $r$ is the radius of the ball, then $h_c\geq r$. 
Larger object size would require larger $h_c$ which increases the size of the basket due to the FoV considerations. Considering these, the camera location is fixed at 0.15 m from the top of the basket. The schematic of how $h_c$ is measured from the top plane of the basket is shown in Fig. \ref{c1-cam-loc}(a). The second reason is that the location of the camera FoV should be free from any obstructions. The scenario is shown in Fig. \ref{c1-cam-loc}(b), where the ideal top and front view of the basket is shown by the figures on the left and right, respectively.  So, in order to ensure a clear view, the minimum basket opening is
\begin{equation*}
    d_{\text{cap}}^{\text{min}}=2h\tan\theta
\end{equation*}
where, $h$ and $\theta$ are the planar depth and FoV angle, respectively, as shown in Fig. \ref{c1-cam-loc}(b).

\subsection{Structural stability of end-effector}
\begin{figure}
        \centering
        \includegraphics[scale=0.35]{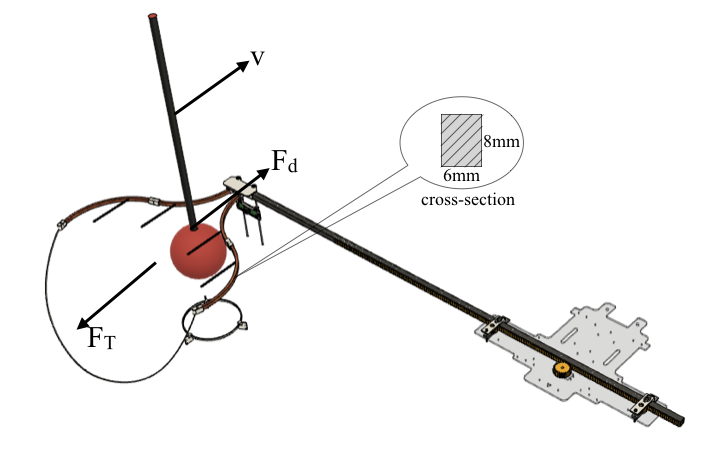}
        \caption{Forces exerted by the end-effector to detach the ball}
        \label{fig:fbd}
\end{figure}

One of the challenges mentioned in Section \ref{sec:2} is the impact on the manipulator and its ability to handle the detachment forces. The maximum drone velocity is $v=6$ m/s. 
\begin{table}
    \centering
    \caption{Total Impact work}
    \begin{tabular}{cc}\hline
        Parameter & Value (J)\\\hline
        $W_{\text {Impact}}$ & 1.08 \\
        $W_{\text {Detach}}$ & 0.06 \\ \hline
        $W_{\text {Total}}$ & 1.14\\\hline
    \end{tabular}
    \label{detach}
\end{table}
Impact strength is calculated in terms of work. Impact work is due to the kinetic energy of the ball. Detachment work is calculated as the detachment force times the detachment distance, which in this case is the diameter of the magnet. The net work is given in Table \ref{detach}. This net work is a representation of the total force. 
Effect of impact is determined using impact strength
\begin{equation*}
    IS = W_{\text {Total}} / A 
\end{equation*}
where, $IS$ is the impact strength and $A$ is the area opposing the impact. Fig. \ref{fig:fbd} shows the forces involved in the detachment process.  The cross-section of the detaching sinusoidal hull is 6 mm $\times$ 8 mm. The impact strength needed to overcome the total work $W_T$ is $IS=23.75$ kJm$^{-2}$. Material selection was based on these impact calculations and experiments performed on different end-effector prototypes.
The final prototype is manufactured using birch wood, which has an impact strength of 92.9 kJm$^{-2}$ \cite{birch}, which ensures adequate strength against the impact. Thus, the end-effector prototype is able to perform effectively with minimum failure. 
\subsection{Manipulator arm}
\renewcommand{\theenumi}{\alph{enumi}}%
\begin{enumerate}
\item Moment due to end-effector: The link joining the end effector and drone is designed as a linear actuated single DoF arm. 
Larger the extension, better is the safety factor. But, stability considerations impose a limit on the maximum extension of the manipulator arm.
The manipulator would act as a cantilever beam with the end-effector and the captured ball acting as an end load. The scenario is shown in Fig. \ref{fig:Arm_cant}. 
\begin{figure}
    \centering
    \includegraphics[scale=0.25]{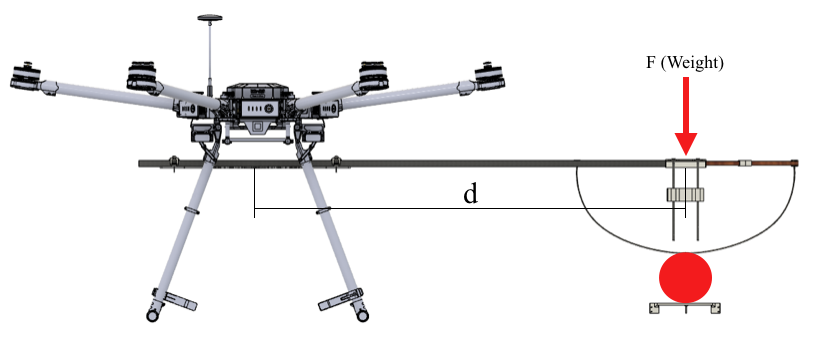}
    \caption{Cantilever loading of the manipulator arm}
    \label{fig:Arm_cant}
\end{figure}
The extension of the manipulator arm for the final design is 1.1 m. The final ball grabber with all accessories weighs 1.4 kg. The corresponding moment at full extension of 1.1 m is 15.107 Nm, which is well within the limits for M600 drone. The extension serves a second purpose. In addition to safety, it also helps to perform the task of grasping without being affected by the downwash. Initial experiments with the smaller drone showed the adverse effect of downwash on the end-effector's effectiveness in grabbing. The ball was often deflected away by the downwash when the drone was close to the ball. By incorporating the extendable arm, this problem is solved. 
\begin{table}
    \centering
    \caption{Relevant parameters for computation of manipulator deflection}
    \begin{tabular}{c c}
    \hline
        Parameter &  Value\\\hline
        F & 13.734 N\\
        I & $2744\times 10^{-12}$ m$^4$\\
        E & 90 GPa \\
        d & 1.1 m\\
        $\delta_{~\text{max}}$ & 0.0246 m\\\hline
    \end{tabular}
    \label{t3}
\end{table}
\item Deflection at the end-effector: The manipulator arm is prone to sag. Deflection calculation is important to account for the error in camera location due to sag. It helps position the camera with the correct orientation. The sag can be calculated for the end-loaded cantilever beam case considered above. The maximum deflection is calculated as
\begin{equation*}
    \delta_{~\text{max}} = \frac{F\times d^3}{3\times E \times I} 
\end{equation*}
where, $\delta_{~\text{max}}$ is the maximum deflection at the free end, $E$ is the modulus of elasticity of the carbon fibre rod \cite{cf} and $I$ is the area moment of inertia due to the carbon fiber tube's cross section. The values for these parameters and the maximum deflection is shown in Table \ref{t3}. The deflection calculation helps in orienting the camera with a slight tilt in the drone's roll axis opposite to the sagging direction to compensate for the error during flight.
\end{enumerate}

\section{Integration and Experimental Results }\label{sec:6}
This section presents the details of the drones used and the associated test rigs for testing the final manipulator prototype, and the results obtained.
\subsection{Test Setup}
\begin{figure}
    \centering
     \includegraphics[scale=0.55]{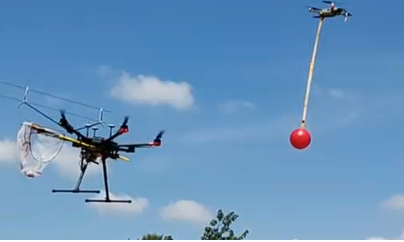}
    \caption{Test setup for stationary and moving ball capture}
    \label{fig:mov-ball}
\end{figure}

Ball grabbing is tested by hanging a red ball of diameter 0.15 m under the DJI Mavic Pro Platinum (Fig. \ref{fig:mov-ball}). The manipulator is tested for stationary drone as well as straight and curved paths of the drone carrying the ball. The control and vision modules for autonomous grabbing are the same as in \cite{tony2020Vision}.

\begin{figure}
    \centering
    \includegraphics[scale=0.65]{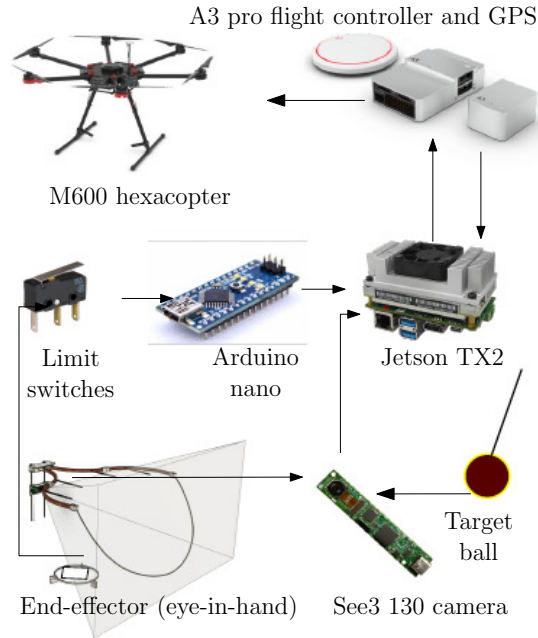}
    \caption{Hardware architecture of the integrated system}
    \label{hard-arch}
\end{figure}

The hardware architecture and the control flow of the integrated system is shown in Fig. \ref{hard-arch}. The avionics and the on-board computers are listed in Table \ref{t2}. 
\begin{table}
    \centering
    \caption{Avionics and on board computers of test drone}
    \begin{tabular}{l c}
    \hline
        Item &  Details\\\hline
        Drone & M600 pro hex-rotor\\
        Auto pilot & A3 pro\\
        Companion board & NVIDIA Jetson TX2\\
        Auxiliary boards & Arduino Mega, Nano\\
        Vision module & See3 130 HD camera\\
        Miscellaneous & Limit switches\\\hline
    \end{tabular}
    \label{t2}
\end{table}
As shown in Fig. \ref{hard-arch}, the ball is detected by the camera fixed at the center of the manipulator end-effector. This information is used by the the TX2 to compute control commands which is sent to the M600 drone via A3 pro flight controller. The localisation is achieved via on-board GPS. When the ball is detached, the limit switches are activated which, via the Arduino nano, sends the information that the ball is captured.

\begin{figure}
    \centering
    \includegraphics[scale=0.5]{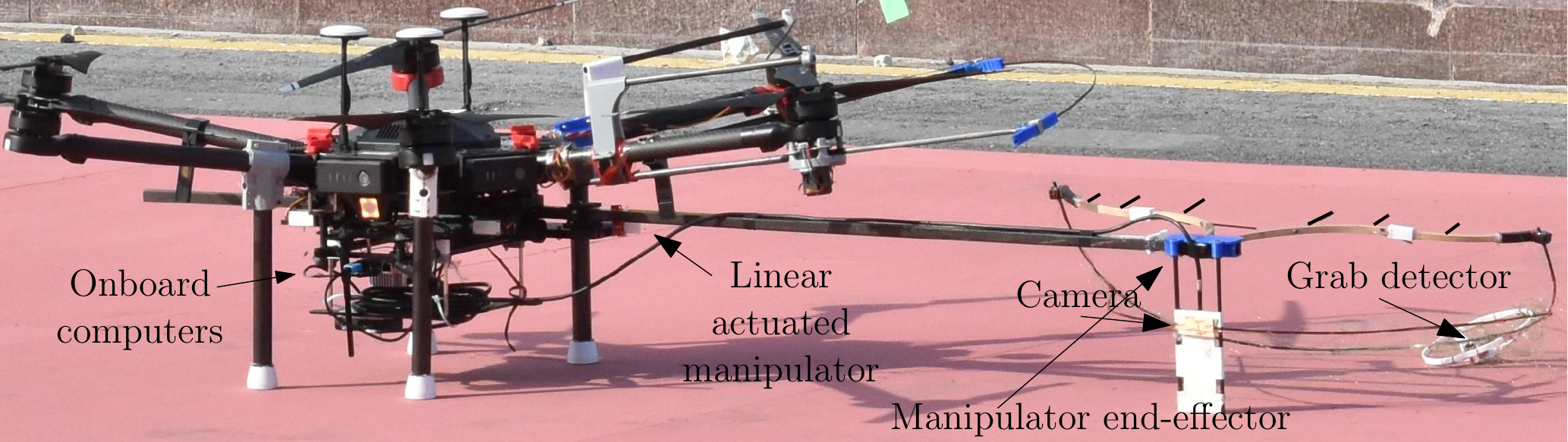}
    \caption{M600 drone integrated with the manipulator}
    \label{fig:drone_grab}
\end{figure}

The final prototype of the proposed manipulation mechanism was tested using DJI M600. The integrated system with the major components labelled, are shown in Fig. \ref{fig:drone_grab}.
\subsection{Field Test Results}
Flight tests are conducted in the test beds using the designed manipulators and drone set up at the airfield of Indian Institute of Science. The environment is windy which tested the robustness of the designed system. The success rate of the manipulator for static ball is found to be 8/10 and maneuvering ball is found to be 7/10. 

\begin{figure}
    \centering
    \begin{subfigure}{1\columnwidth}
        \centering
        \includegraphics[scale=0.45]{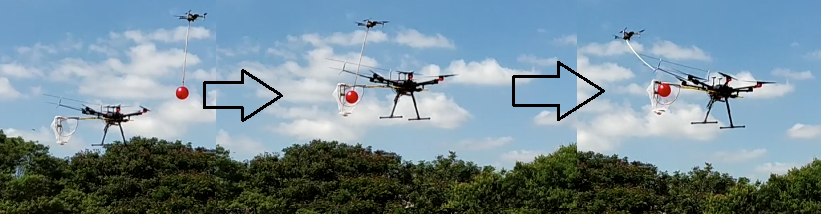}
        \subcaption{}
    \end{subfigure}
    \begin{subfigure}{1\columnwidth}
        \centering
        \includegraphics[scale=0.55]{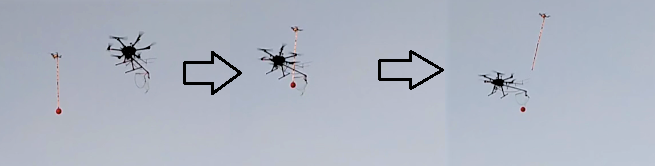}
        \subcaption{}
    \end{subfigure}
    \begin{subfigure}{1\columnwidth}
    \centering
    \includegraphics[scale=0.5]{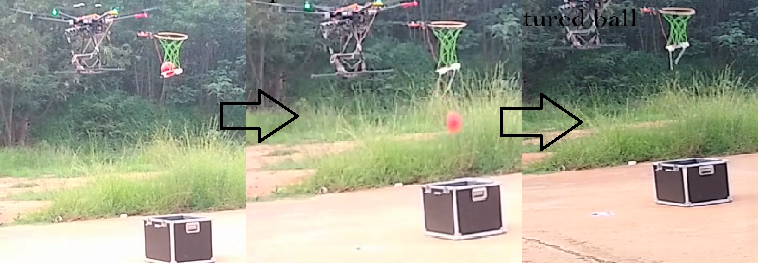}
    \subcaption{}
    \label{fig:ball-drop}
\end{subfigure}
    \caption{Snapshots of grabbing of (a) static ball (b) moving ball (c) Snap shot of ball dropping}
    \label{fig:mov-ball-grab}
\end{figure}

A snapshot of a grabbing instance for static ball is shown in Fig. \ref{fig:mov-ball-grab}(a), where the manipulator moves towards the ball and pulls it to detach from the magnetic attachment. Instants of moving ball capture are shown in Fig. \ref{fig:mov-ball-grab}(b), where the ball is detached from a moving drone and the captured ball is collected in the mesh under the basket. The dropping exercise using the release mechanism at the bottom of the basket end-effector is shown in Fig. \ref{fig:ball-drop}. The drone approaches the box in which the ball is to be deposited, followed by actuating the servo to release the ball. The videos of the experimental results from which the above instants are captured, can be found in \footnotemark[1]. 
\footnotetext[1]{\url{https://youtu.be/1jdtIumUvdI}}
\subsection{Discussion}

Some interesting observations were made during the experiments. The visual feedback, grabbing algorithm, system dynamics, and the inherent delays in computation, were found to have an impact on the success rate. This points to the unavoidable coupling between the manipulator design and the software used to perform the mission. There is a fine relationship between them which decides the success rate.
The proper location of the camera was also found to be an important parameter in successful grabbing, considering the field of view and approach direction. Effects of wind gust is minimal for the proposed design, but any disturbance above 12 m/s brings in some vibrations. Sag of the manipulator arm was also observed over time, which was primarily due to the weight of wires at the end-effector side. Nevertheless, proper calibration of sensors resulted in good success rate. With the present design, any object within 0.15 kg and within 0.2 m diameter could be grabbed successfully. 

Few design and manufacturing aspects in the final design were compromised to fulfil the requirements, which could be improved in the future variants. The shape of the top portion was an intuitive design and gave good success rates, but a deeper analysis would provide more effective shapes for similar end-effector footprint. The detachment structures and their locations could be analysed for better performance.   

\section{Conclusions}\label{sec:7}
This work provided design, development and testing details of an aerial manipulation for grasping of dynamic targets. The problem statement was inspired by Challenge 1 of MBZIRC 2020. The major complexities involved in the design and development are discussed in detail. The conceptual design for the task was presented, describing the reasons for design and the material choice. The experimental results and relevant observations are also reported in this paper. The obvious and unavoidable coupling between the hardware design and software modules are pointed out. The repeatability and success rates of the final configurations are reported and possible developments on the final design are also presented. The manipulator has been designed for a ball grabbing in this work but can be modified for many other applications.

\section*{Acknowledgements}
 We acknowledge members of GCDSL for their valuable suggestions in development of the manipulator. The contributions of Integrative Multiscale Engineering Materials and Systems (iMEMS) lab and Advanced Materials and Processing Laboratory (AMPL), especially Sagar K. and Abhishek Nagaraja, towards prototyping manipulator components, deserves thankful mention. We acknowledge our collaborator, Tata Consultancy Services for their contributions towards this work.
 
 \section*{Author Contributions}
Author A designed the manipulator. Authors AB prototyped the manipulator and wrote the paper. Author C was the pilot for the tests and dealt with the hardware integration. Authors DEF carried out the software integration to automate the process and helped with the tests. Author G conceived the design idea and wrote the paper. Author F lead the project, reviewed work progress and wrote the paper.

\section*{Financial Support}
This work is partially supported by Robert Bosch Center for Cyber Physical Systems, (IISc) and Khalifa University, Abu Dhabi, UAE.
\section*{Competing Interest Declaration}
Competing interests: The author(s) declare none.

\end{document}